\documentclass{article}

\usepackage{microtype}
\usepackage{graphicx}
\usepackage{subfigure}
\usepackage{booktabs} 

\usepackage{hyperref}

\usepackage[accepted]{icml2023}

\usepackage{amsmath}
\usepackage{amssymb}
\usepackage{mathtools}
\usepackage{amsthm}

\usepackage[capitalize,noabbrev]{cleveref}

\theoremstyle{plain}

\theoremstyle{definition}

\theoremstyle{remark}

\usepackage[textsize=tiny]{todonotes}

\icmltitlerunning{MOMA: Distill from Self-Supervised Teachers}

\begin{document}

\twocolumn[
\icmltitle{MOMA: Distill from Self-Supervised Teachers
}



\icmlsetsymbol{equal}{*}

\begin{icmlauthorlist}
\icmlauthor{Yuchong Yao}{}
\icmlauthor{Nandakishor Nandakishor}{}
\icmlauthor{Marimuthu Palaniswami}{}
\end{icmlauthorlist}


\icmlcorrespondingauthor{Yuchong Yao}{yuchongy1@student.unimelb.edu.au}

\icmlkeywords{Self-Supervised Learning, Knowledge Distillation, Masked Image Modelling, Contrastive Learning, Vision Transformer, Deep Learning, Machine Learning}

\vskip 0.3in
]




\begin{abstract}
Contrastive Learning and Masked Image Modelling have demonstrated exceptional performance on self-supervised representation learning, where Momentum Contrast (i.e., MoCo) and Masked AutoEncoder (i.e., MAE) are the state-of-the-art, respectively. In this work, we propose MOMA to distill from pre-trained \underline{MO}Co and \underline{MA}E in a self-supervised manner to collaborate the knowledge from both paradigms. During the distillation, the teacher and the student are fed with original inputs and masked inputs, respectively. The learning is enabled by aligning the normalized representations from the teacher and the projected representations from the student. This simple design leads to efficient computation with extremely high mask ratio and dramatically reduced training epochs, and does not require extra considerations on the distillation target. The experiments show MOMA delivers compact student models with comparable performance to existing state-of-the-art methods, combining the power of both self-supervised learning paradigms. It presents competitive results against different benchmarks in computer vision. We hope our method provides an insight on transferring and adapting the knowledge from large-scale pre-trained models in a computationally efficient way.
\end{abstract}

\section{Introduction}

Self-supervised learning (SSL) \cite{he2020momentum}\cite{he2022masked} has shown impressive potential in various vision tasks and applications, owing to increasingly available data and advancing hardware. SSL extracts semantically rich information from large-scale unlabelled data and delivers a foundation model (e.g., \cite{bao2021beit}, \cite{devlin2018bert}) whose representations can be transferred for downstream tasks. Among the blossom of self-supervised learning methods, there are two dominant branches: contrastive learning and masked image modelling. 

Contrastive learning (e.g., \cite{chen2020simple}, \cite{he2020momentum}) enables the unsupervised learning by maximizing the agreement between two different augmented views from the same input. The key is to introduce reliable and challenging data augmentations that encourages semantically meaningful representations. Contrastive learning approaches have demonstrated exceptional results in the past few years, which even surpassed supervised learning algorithms. Recently, masked image modelling (e.g., \cite{xie2022simmim}, \cite{he2022masked}) has become another main paradigm for learning self-supervised vision representations. The idea of masked image modelling stems from the success of masked language pre-training (e.g., \cite{devlin2018bert}, \cite{brown2020language}) in Natural Language Processing. The objective is to reconstruct original images from partially masked inputs. Masked image modelling presents high efficiency under a high mask ratio and achieves superior performance than contrastive learning across various benchmarks (e.g., \cite{deng2009imagenet} \cite{lin2014microsoft}).

However, both contrastive learning and masked image modelling suffer from their own limitations. Contrastive learning relies heavily on data augmentations and requires additional techniques such as memory bank \cite{wu2018unsupervised}, momentum encoder \cite{he2020momentum}, and stop-gradient \cite{chen2021exploring}. In \cite{chen2021empirical}, the authors also pointed out the necessity to freeze the patch embedding layer when training with vision transformers \cite{dosovitskiy2020image}. Additionally, the quality of negative samples \cite{robinson2020contrastive} is also critical for contrastive learning. As for masked image modelling, it optimizes a pixel-level objective, which gains low-level representation and knowledge from the images. Therefore, such pre-training lacks of high-level representation, especially the semantic meanings behind the images.

Recent studies \cite{chung2021w2v} \cite{huang2022contrastive} \cite{mishra2022simple} \cite{zhou2022mimco} \cite{yao2022masked} attempt to combine the power of contrastive learning and masked modelling, yielding promising results. They suggest that both paradigms are complementary with each other and can deliver stronger representations when they are combined into a unified framework. Furthermore, integrating two paradigms into one framework introduces higher computational cost, which requires extensive resources (e.g., hundreds of GPU hours, enormous memory capacity, and excessive storage requirements). It is also not energy-efficient to training different frameworks from the scratch as they all tend to require large training epochs but the resulting difference in the performance is often negligible.

In this work, we introduce \textbf{MOMA}, which integrates knowledge from pre-trained contrastive learning (i.e., \textbf{MO}co) and masked image modelling (i.e., \textbf{MA}sked autoencoder) through knowldge distillaiton \cite{hinton2015distilling}. There are three options presented in MOMA: (1) Distil from pre-trained MoCo to pre-trained MAE. (2) Distil from pre-trained MAE to pre-trained MoCo. (3) Distil from both pre-trained MoCo and MAE to a random initialized student model. We feed the original image to the teacher model and pass masked or intensively augmented samples to the student model. The learning objective is straightforward, which aligns the representations from normalized teacher outputs and reconstructed student outputs. This design leads to a simple and efficient framework for combining both contrastive learning and masked modelling. MOMA can accept an extremely high mask ratio during training, which leads to lower computational cost and faster training speed. Instead of training from the scratch, MOMA fully uses the pre-trained checkpoints from existing state-of-the-art paradigms. It enables MOMA to achieve excellent performance within only limited number of training epochs, which saves computation,energy and achieves competitive performance across different tasks. Additionally, it does not require a sophisticated design for knowledge distillation objectives as it directly aligns the representations from teacher and student. Finally, MOMA makes it possible to extract a more compact and lightweight model that fuses the power of different self-supervised learning paradigms. The proposed work enables new framework and mechanisms to utilize large-scale self-supervised models effectively and perform  transfer in an energy-efficient manner.

\section{Related Work}

\begin{figure}[H]
\vskip 0.2in
  \centering
  \includegraphics[width=0.35\textwidth,height=4.5cm]{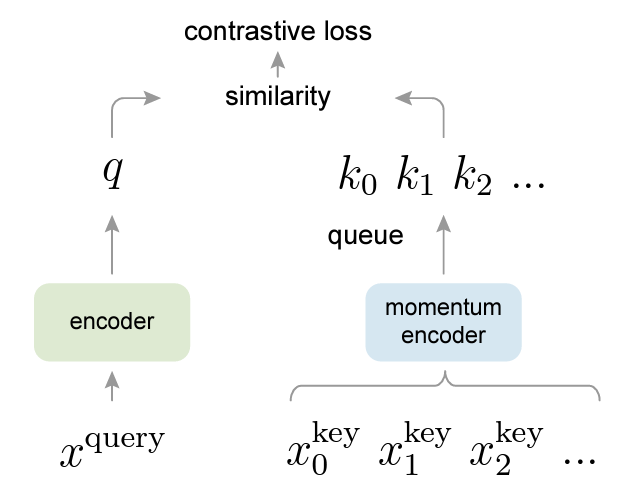}
   \caption{\textbf{Momentum Contrast (MoCo).} MoCo \protect\cite{he2020momentum} is one of the state-of-the-art method in contrastive learning. It proposed a siamese network structure, where the overall objective is to minimize the contrastive loss (maximizing the agreement) between the main encoder and the momentum encoder.}
   \label{fig:moco}
   \vskip -0.2in
\end{figure}

\textbf{Contrastive Learning.} This branch of self-supervised learning approaches stems from the idea of instance discrimination \cite{wu2018unsupervised}, which treats each sample as a class. Each sample goes through strong data augmentation operations. The augmented views from the same instance are positive pairs, whereas the augmented views from different instances are negative pairs. The learning is enabled by maximizing the agreement between positive samples (or disagreement between negative pairs). Memory bank is one of the critical components in this process, which ensures the diversity of negative samples. MoCo \cite{he2020momentum} (see \textbf{Figure \ref{fig:moco}}) introduced a momentum component (updated by exponential moving average) in the framework, which further improves the performance of siamese learning network. SimCLR \cite{chen2020simple} showed that non-linear projector, large batch size and stronger combinations of data augmentation operations are critical for the performance in contrastive learning. Later, the improved version of MoCo \cite{chen2020improved} \cite{chen2021empirical} and SimCLR \cite{chen2020big} further improved the benchmarks by integrating each other's techniques and adopting larger vision transformers \cite{dosovitskiy2020image}. SwAV \cite{caron2020unsupervised} introduced clustering into the framework, which eases the requirements for large number of negative samples. BYOL \cite{grill2020bootstrap} applied an additional predictor after the projector in the contrastive learning framework and showed that the method could achieve excellent results without negative samples. DINO \cite{caron2021emerging} further improved the BYOL's idea and incorporated self-distillation. Simsiam \cite{chen2021exploring} introduced stop-gradient operation, demonstrating that prevention of the mode collapse is the essential part of contrastive learning.

\begin{figure}[H]
\vskip 0.2in
  \centering
  \includegraphics[width=0.4\textwidth,height=5cm]{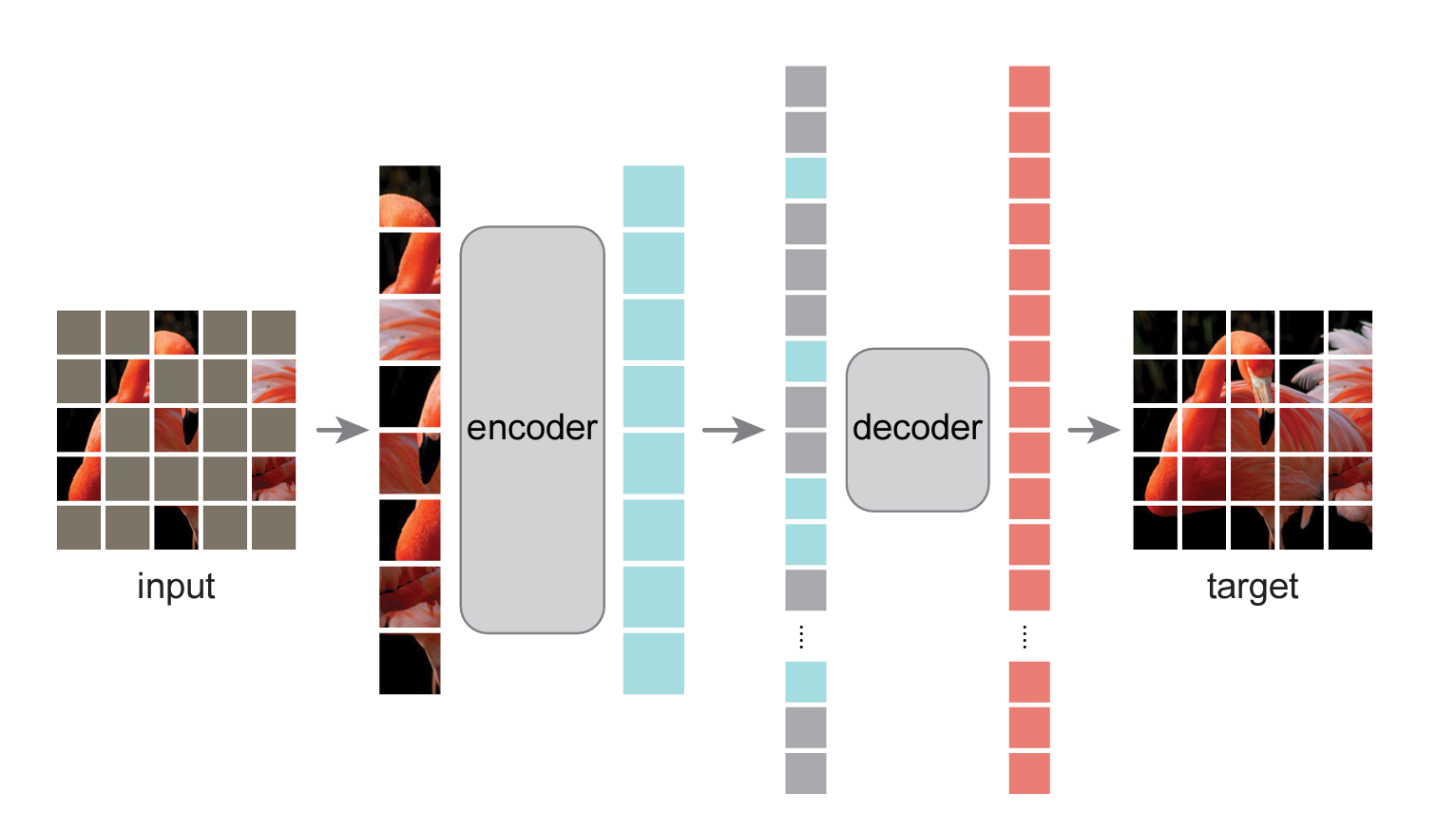}
   \caption{\textbf{Masked AutoEncoder (MAE).} MAE \protect\cite{he2022masked} is a powerful masked image modelling method which was proposed recently. It masks large portion of the image and applies asymmetric encoder-decoder network to restruct the original image. Its simple design leads to fast and effective self-supervised representation learning}
   \label{fig:mae}
   \vskip -0.2in
\end{figure}

\begin{figure*}[t]
\vskip 0.2in
  \centering
  \includegraphics[width=0.75\textwidth,height=5cm]{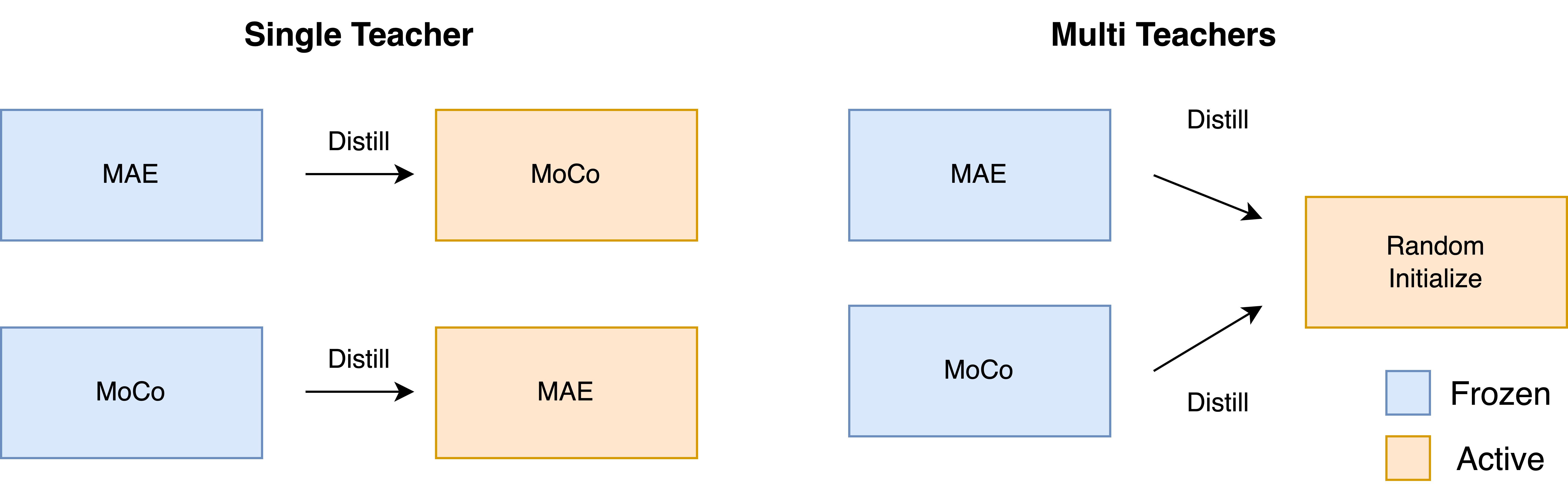}
   \caption{\textbf{Overview of MOMA.} During distillation, the teacher model is frozen without gradient update and the student is updated through gradient. In the single teacher setting, the knowldge is either distilled from a MAE to a MoCo or the reverse direction. In the multiple teachers setting, knowledge from MoCo and MAE is distilled to a randomly initialized student.}
   \label{fig:moma_pic}
   \vskip -0.2in
\end{figure*}

\textbf{Masked Image Modelling.} The early work \cite{pathak2016context} introduced inpainting as a pretext task for self-supervised learning, which reconstructs corrupted inputs for self-supervised learning. iGPT \cite{chen2020generative} performed reconstruction on corrupted images, following the auto-regressive approach described in GPT \cite{brown2020language}. In contrast, BEiT \cite{bao2021beit} followed a BERT \cite{devlin2018bert} style pre-training paradigm, which recovered the masked images tokens in an autoencoding manner. It adopts a pre-trained tokenizer in the framework to transform input images into visual tokens. MAE \cite{he2022masked} (see \textbf{Figure \ref{fig:mae}}) and SimMIM \cite{xie2022simmim} are two concurrent works that present an end-to-end framework with asymmetric encoder-decoder architecture, and they adopted a high mask ratio to boost the computational efficiency. The idea is simple and straightforward, where the masking strategy can be as simple as random masking. In MAE, the encoder took unmasked patches and the decoder reconstructed the original images based on encoded visible tokens and masked tokens. The result surpassed the previous state-of-the-arts presented by contrastive learning. Moreover, the latter supports both vision transformers \cite{dosovitskiy2020image} and hierarchical vision transformers (Swin \cite{liu2021swin}). MaskFeat \cite{wei2022masked} applied masks on the histogram of gradient (HOG) features and reconstructed those features to enable learning.

\textbf{Combining Two Paradigms.} Recent work attempted to combine the power of contrastive learning and masked image modelling. SIM \cite{tao2022siamese} incorporated masking as part of the data augmentation operations into the contrastive learning framework. iBOT \cite{zhou2021ibot} adopted a siamese network structure as contrastive learning and minimize the distance between the masked branch and the unmasked branch. MimCo \cite{zhou2022mimco} tried to improve the linear separability of masked image modelling by introducing a two-stage pre-training methods that includes contrastive learning and masked image modelling. CAN \cite{mishra2022simple} applied mask on both branches in siamese network and optimized an InfoNCE loss \cite{oord2018representation}, a reconstruction loss, and a denoising loss. CMAE \cite{huang2022contrastive} computed a reconstruction loss and a contrastive loss based on the decoder's outputs between the online branch and the target branch. MACRL \cite{yao2022masked} proposed an asymmetric siamese network structure, which applied masks on the online branch and passed the original images into the target momentum branch (both encoder and the projector). The framework optimized a contrastive objective based on the encoded representations and a reconstruction objective based on the decoded outputs.

\textbf{Knowledge Distillation.} The idea is proposed in \cite{hinton2015distilling}, which introduced a way to transfer knowledge from a well-trained teacher model to a more compact or compressed student model. Existing methods also attempted to utilize knowledge distillation as part of the self-supervised learning framework. MVP \cite{wei2022mvp} and MaskDistill \cite{peng2022unified} include a CLIP \cite{radford2021learning} as the teacher to guide self-supervise learned features. Teacher with high capacity and rich representation leads to stronger students. dMAE \cite{bai2022masked} presented masked knowledge distillation on the intermediate features between student and teacher. The method optimized a mask reconstruction loss and a $L_1$ distillation loss. dBOT \cite{liu2022exploring} introduced multi-stage masked knowledge distillation from itself. Between stages, the teacher is re-initialized with the student weights with exponential moving average and student is randomly initialized. With bootstrapped teachers, dBOT can achieve better performance than the existing baseline. 

\begin{figure*}[t]
\vskip 0.2in
  \centering
  \includegraphics[width=0.7\textwidth,height=5.5cm]{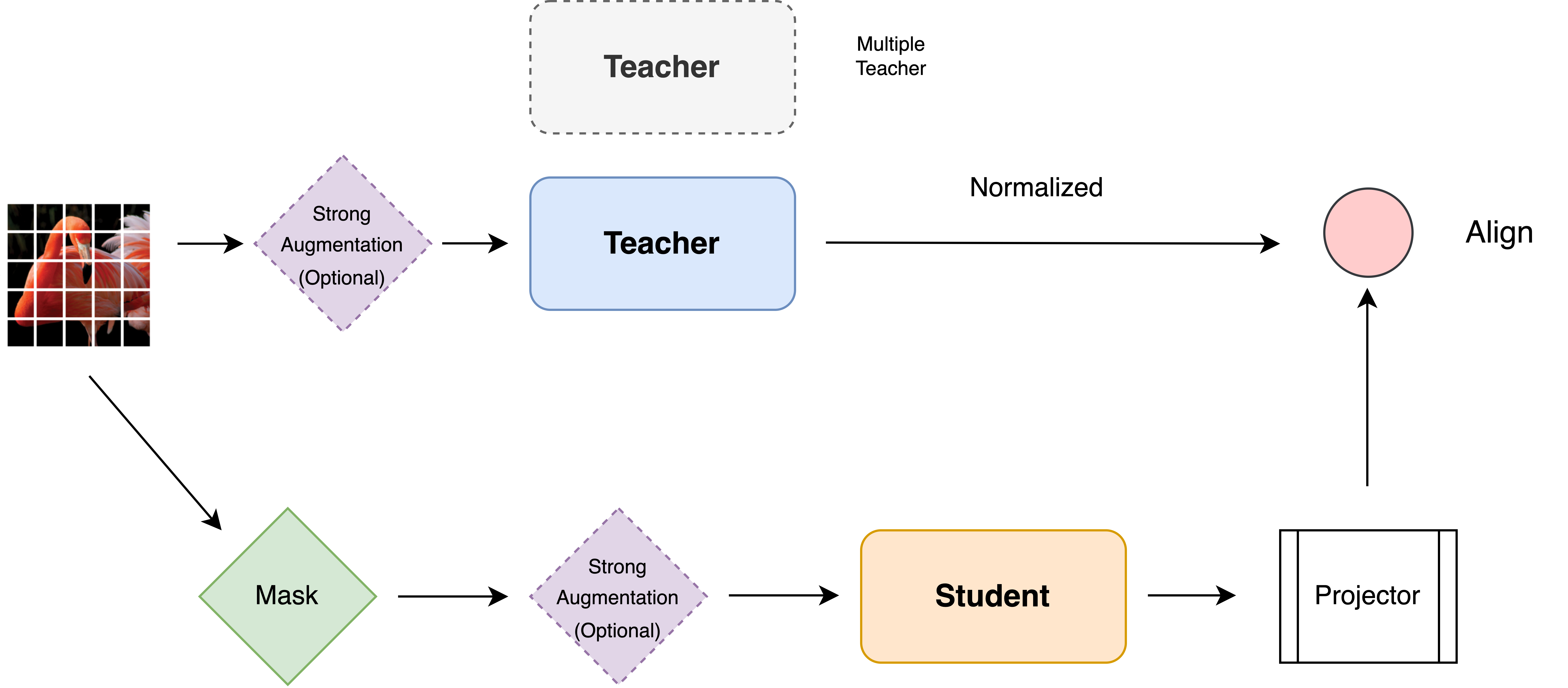}
   \caption{\textbf{Distillation Procedure of MOMA.} MOMA follows asymmetric siamese network structure. The teacher takes the original image (or optionally strong augmented image) and the student takes masked image (or with optionally extra strong data augmentations). The objetive is to align the outputs between normalized teacher representation and projected student representation.}
   \label{fig:moma_detail}
   \vskip -0.2in
\end{figure*}

\section{Approach}

\subsection{Preliminary}

\textbf{Momentum Contrast} Our work borrows some of the insights from the state-of-the-art contrastive learning method MoCo v3 \cite{chen2021empirical}. The ideas of MoCo is shown below: 

\begin{equation}
\mathcal{L}_{MoCo}=-\log \frac{\exp \left(q \cdot k_{+} / \tau\right)}{\sum_{i=0}^K \exp \left(q \cdot k_i / \tau\right)}
    \label{eq:moco}
\end{equation}

where $q$ is encoded query from the main encoder and $k$ is the encoded key from the other branch's momentum encoder. The framework minimizes the $\mathcal{L}_{MoCo}$ where query is supposed to be similar to positive keys (i.e., $k_{+}$) and dissimilar to all other (negative) keys. The siamese network architecture of teacher and student works in the similar way as the momentum setup in MoCo, where there is a frozen teacher branch and an updating student branch. However, the objective is to minimize the Smooth $L_1$ loss between the two branches rather than the traditional contrastive loss \cite{oord2018representation}. Moreover, we utilize the strong data augmentation operations mentioned in MoCo, including combinations of  Gaussian blurring, solarization, colour jittering, and grey scaling. Such augmentation methods encourage model to learn invariant features that are more robust and generalizable. Furthermore, we adopt large batch size (i.e., 4096) as suggested by MoCo, enabling diverse feature sets and representations learning.

\textbf{Masked AutoEncoder} Masked AutoEncoder \cite{he2022masked} is the dominant approach in visual pre-training, surpassing the performance of contrastive learning with less computational requirements. MAE adopts asymmetric encoder-decoder architecture, where the encoder takes visible tokens (randomly masked out from the original inputs) and the decoder processes the encoded representation and the masked tokens to reconstruct the original inputs. The objective is shown below:
\begin{equation}
\mathcal{L}_{MAE}=\mathcal{L}\left(\mathcal{D}_\theta \circ \mathcal{E}_\theta \left(\mathbf{x} \odot \mathcal{M}\right), \mathbf{x}\right)
\label{eq:mim}
\end{equation}
where $\mathcal{E}$ and $\mathcal{D}$ are the encoder and decoder, respectively. $\mathcal{M}$ stands for the random mask applied on the input $x$.
It applies random masking with high mask ratio (e.g., 75\%) which saves huge computation for its encoder. In our work, we apply random masking to the inputs as well, with similar or even higher mask ratio. Additionally, the mask is not limited to the student branch, but also eligible for the teacher(s) branch. By applying masks for both teachers and student, we enable efficient training for multiple teachers and scale our framework effectively. Different from the original MAE, our target is not to reconstruct the pixel values of original images, but to align the features for the teacher(s) and student. Therefore, we don't have the decoder component for the teacher and student, which further saves the computation. Furthermore, feature alignment encourages the model to maintain high level semantical inforamtion rather than pixel level reconstruction.

\subsection{Overview of MOMA}

\textbf{Teacher.} We utilized publicly available checkpoints from pre-trained MoCo \cite{chen2021empirical} and MAE \cite{he2022masked}. We used the ViT-Base model, which is a 12-layer vision transformer \cite{dosovitskiy2020image} with 12 attention heads for most of our experiments. We also use the ViT-Large model (i.e., 24-layer,16 heads, and 1024 embedding size) to further boost the performance. There are three different settings for the teacher. In the single-teacher setup, we either used a pre-trained MoCo model or a pre-trained MAE as the teacher. We utilized both pre-trained MoCo and pre-trained MAE as teachers for the multi-teacher setup. The weights are frozen for all layers in the teacher model.

\textbf{Student.} We adopted different settings for single-teacher and multi-teacher setups.In single-teacher distillation, a pre-trained MAE is the student model when the pre-trained MoCo is the teacher. On the other hand, a pre-trained MoCo will be treated as the student if the pre-trained MAE is the teacher model. Therefore, in the single-teacher setup, the student model has the sizes limited by those pre-trained checkpoints (i.e., ViT-Base). In the multi-teacher setup, we used a random initialized vision transformer as the student. As the student is randomly initialized, we can use model of arbitrary size. By default, we adopted a 12-layer 12-head vision transformer for fair comparison with the single-teacher setup. In practice, we are able to use more compact and lightweight student for more efficient processing and storage, such as ViT-Small (i.e., 12-layer, 12-head, 384 embedding dim). \textbf{Figure \ref{fig:moma_pic}} provides an overview of MOMA framework and illustrates the single-teacher and multiple-teacher setups.

\textbf{Distillation Procedure.} For the teacher(s), a normalization layer will be applied after the inputs go through the transformer layers. We incorporated Layer Normalization \cite{ba2016layer} to produce the normalized features as it is the de-facto choice for vision transformers. For the student model, there is a non-linear projection head following the transformer layers. The projection head is a fast and lightweight single linear layer, which projects the learned features from the student, and matches the dimension with the teacher. During training, only the student model is active and the teacher model is frozen without any gradient update. The knowledge distillation is achieved by aligning the outputs from teacher(s) and the student branches. We applied a Smooth $L_1$ loss for training the framework. In practice, we can distill from a larger teacher model (e.g., ViT-Base, ViT-Large) to a smaller student model (e.g., ViT-Base, ViT-Small) to achieve good performance and computational efficiency. The distillation procedure of MOMA is displayed in \textbf{Figure \ref{fig:moma_detail}}.

\subsection{Single Teacher}

The general procedure of single teacher distillation is shown below:

\begin{equation}
\mathcal{L}_{Single}={Smooth}\mathcal{L}_{1}\left(\mathcal{P} \circ \mathcal{S}\left(\mathbf{x} \odot \mathcal{M}\right),   \mathbf{N}\circ\mathcal{T}\left(\mathbf{x}\right)\right)
\label{eq:single}
\end{equation} where $\mathcal{N}$ is the normalization, $\mathcal{P}$ is the projector, $\mathcal{M}$ denotes the mask. $\mathcal{T} and \mathcal{S}$ refer to the teacher and student, respectively. If the teacher or student takes samples with strong data augmentations, we can replace $\mathbf{x}$ with $\mathbf{x}_{aug}$ in the above formula.

\textbf{MoCo to MAE.} In this setup, a pre-trained MoCo model is treated as the teacher  and a pre-trained MAE model is treated as a student. During the forward pass, original images are fed into the teacher model, which go through the attention blocks and the normalization layer. Masked images are fed into the student, saving the computational cost (i.e., smaller portion of total inputs) and making the learning more challenging. The masked ratio can be even higher than the original configuration described in MAE \cite{he2022masked} (e.g., 75\%). There is another option to feed in strong augmented samples (as described in MoCo\cite{chen2021empirical}) into the student model. The reason for this setup is that we want to force the student to perform the learning task which its teacher was previously pre-trained on (i.e., MoCo is pre-trained on strong augmented samples). 

\textbf{MAE to MoCo.} This is the reverse setup compared to the "MoCO to MAE" settings, where the teacher is a pre-trained MAE and the student is a pre-trained MoCo. The teacher model processes the original images and the student model takes the masked inputs with high mask ratio. Since the teacher model's pre-training is masked image modelling already (no strong augmentation used in teacher's pre-training), we did not apply strong augmentation for the student model.

\subsection{Multiple Teachers.}

The distillation procedure for multiple teachers is shown as follows:

\begin{equation}
\begin{split}
\mathcal{L}_{Multi} = \alpha \times  {Smooth}\mathcal{L}_{1}\left(\mathcal{P} \circ \mathcal{S}\left(\mathbf{x} \odot \mathcal{M}\right),   \mathbf{N}\circ\mathcal{T}_{MAE}\left(\mathbf{x}\right)\right) \\ + \beta \times {Smooth}\mathcal{L}_{1}\left(\mathcal{P} \circ \mathcal{S}\left(\mathbf{x} \odot \mathcal{M}\right),   \mathbf{N}\circ\mathcal{T}_{MoCo}\left(\mathbf{x}\right)\right)
\label{eq:mul}
\end{split}
\end{equation}

where $\mathcal{T}_{MAE}$ and $\mathcal{T}_{MoCo}$ denote the MAE teacher and MoCo teacher, respectively. The remaining notations follow the same convention as the single teacher setup.

In this setup, we adopted two pre-trained teacher models: a MoCo and a MAE. We set the student model to be a randomly initialized vision transformer, but it is also possible to use a pre-trained student model. During the distillation, original images are passed to the teacher models and their own normalization layers. Two extracted representations are obtained from the teachers. As for the student model, we feed two sets of inputs: strong augmented samples, and masked samples. The design consideration follows the idea of the single-teacher setup, where we encourage the student model to perform the same learning task as its teacher(s) was previously pre-trained on. Therefore, we also obtain two extracted representations from the student model. The learning objective is to minimize the distance between the representations from the teachers and the student, which is shown in \textbf{Equation}. We distill from two ViT-Base teachers into a ViT-Base student. In practice, we can use more lightweight student model such as ViT-Small.

\section{Experiments}

\subsection{ImageNet-1K Results}

We perform the knowledge distillation as a self-supervised pre-training on ImageNet-1K \cite{deng2009imagenet} dataset. It contains 1.2 million images, where the input size is set to $224 \times 224$ and there are 1,000 classes in total (class information is not used during the knowledge distillation / pre-training). We utilized patch size of 16 for ViT-Small (12-layer/6-head), ViT-Base (12-layer/12-head), and ViT-large (24-layer/16-head) model; therefore, the total token size is 196.

\textbf{Batch Size.} Although we utilized the pre-trained weights from MoCo v3 \cite{chen2021empirical}, our framework is not based on contrastive learning. Therefore, there is no need for large batch size. According to the setup in MoCo v3 and MAE \cite{he2022masked}, both adopted a batch size of 4,096. Therefore, we use batch size 4,096 by default during the pre-training stage. For fine-tuning, we adopted a batch size of 1,024.

\textbf{Data Augmentation.} During pre-training, we applied random resize crop and random horizontal clip for the original samples and the randomly masked samples. For the strong augmented samples, we followed the procedure described in MoCo v3 \cite{chen2021empirical} and BYOL \cite{grill2020bootstrap}, where additional color jittering, grey scaling, gaussian blurring, and solarization were adopted. In fine-tuning, we utilized random resize crop, Autoaugment \cite{cubuk2018autoaugment}, and Cutmix \cite{devries2017improved}.

\textbf{Optimizer.} Following the setup in MoCo v3 \cite{chen2021empirical} and MAE \cite{he2022masked}, we adopted AdamW \cite{loshchilov2017decoupled} for pre-training, fine-tuning. A cosine annealing scheduler \cite{loshchilov2016sgdr} is applied for all optimizers with 5 warm up epochs.  

\textbf{Pre-training.} By default, we train the distillation framework in the pre-training stage for 100 epochs with 20 warm up epochs. We adopt a learning rate of 1.5e-4, weight decay rate of 0.05, and $\beta_1, \beta_2$ of 0.9, 0.95. We employed a extremely high mask ratio of 90\%.

\textbf{Fine-tuning.} The weight obtained from the pre-training stage is tuned end-to-end with the ground-truth labels. We adopted a learning rate of 1.5e-3, a weight decay rate of 0.05, and $\beta_1, \beta_2$ of 0.9, 0.999. We trained for 100 epochs with 5 warm up epochs.

\textbf{Main Results.}  As we can see from the \textbf{Table \ref{tab:image1kft}}, our methods outperform the existing self-supervised learning approaches on ImageNet. It adopts a ViT-Large from MAE as teacher and a ViT-Base from MoCo as the student. Comparing to those methods, MOMA utilizes much less training epochs, which is much more computationally efficient and energy-saving. Moreover, it adopted extremely high mask ratio (i.e., 90\%) to further boost the performance and efficiency, which is much higher than the original MAE setup (i.e., 75\%). Furthermore, it show better results than the supervised training counterpart, which is trained for 300 epochs.

\begin{table}
    \centering
    \caption{\textbf{Fine-tuning accuracy on ImageNet-1K.} All methods use ViT-Base model.}
    \vskip 0.15in
    \begin{tabular}{lrr}
        \toprule
        Method  &  Epochs & Top-1 Acc (\%)\\
        \midrule
        Sup. \cite{touvron2021training}    & 300         & 81.8        \\
        MAE  \cite{he2022masked}      & 1600         & 83.6         \\
        MoCo v3  \cite{chen2021empirical} & 300          & 83.2       \\
        SimMIM \cite{xie2022simmim}  & 800          & 83.8      \\
        BEiT \cite{bao2021beit}  & 800 & 83.2 \\
        MOMA & 100 & 84.2 \\
        \bottomrule
    \end{tabular}
    
    \label{tab:image1kft}
    \vskip -0.1in
\end{table}

\subsection{Ablation Study}

\textbf{Distillation Options.} We evaluate the effect of different distillation methods. As we can see from \textbf{Table \ref{tab:disopt}}, using a MoCo pre-trained model as the teacher and a MAE pre-trained model as the student yields the best results. Comparing with distilling from MAE to MoCo, MoCo teacher achieves better results because our proposed distillation approach is built upon masked image modelling. Therefore, a contrastive learning-based teacher helps compensate for the semantic and high level knowledge. Still, Masked image modelling teacher advances the overall performance. The reason is that our framework performs masked modelling on feature-level semantic information (i.e., align the representations from teacher and student), rather than pixel-level construction. Therefore, our propose method enjoys both low-level and high-level knowledge. However, as the model scales, MAE pre-trained model shows significant better finetune performance over MoCo pre-trained model. Therefore, for teacher of size ViT-Large, we adopted MAE pre-trained models as the teacher for better performance. As for the multiple teachers setup, it does not perform as good as the single teacher setup. Moreover, multiple teachers increases the computational burden during the pre-training. The current strategy for learning from multiple teacher is averaging the alignment loss from both teachers. We can possibly improve the performance by applying a weighted average with learnable parameters. It is preferred less than the single-teacher setup due to the computation trade-off.

\begin{table}

    \centering
    
    \caption{\textbf{Comparison of fine-tuning accuracy on ImageNet-1K from different distillation options.} All methods use ViT-Base model unless otherwise stated. Both teacher (s) and student are ViT-Base model. Mask ratio is 90\% for all entries.}
    \vskip 0.15in
    \begin{tabular}{lr}
        \toprule
        Method   & Top-1 Acc (\%)\\
        \midrule
        MAE to MoCo  & 83.7 \\
        MoCo to MAE  & 84.0 \\
        Multi   & 83.4 \\
        \bottomrule
    \end{tabular}
    
    \label{tab:disopt}
    \vskip -0.1in
\end{table}

\textbf{Model Size.} We estimate how the size of teacher model and student model affect the performance. As shown in \textbf{Table \ref{tab:dissize}}, all teacher models are MAE pre-trained and all student models are MoCo pre-trained. We can see that larger teacher and larger student lead to better performance, which is understandable as larger models have larger capacity. Furthermore, small student can achieve comparable performance after distillation from the larger teacher, leading to parameter-efficient models. The lightweight student models saves storage and computation power, with acceptable trade-off over the performance.

\begin{table}
    \centering
    \caption{\textbf{Comparison of fine-tuning accuracy on ImageNet-1K for different model size.} All methods use ViT-Base model unless otherwise stated. All teacher models are MAE pre-trained, all students are MoCo pre-trained. Mask ratio is 90\% for all entries.}
    \vskip 0.15in
    \begin{tabular}{lr}
        \toprule
        Method   & Top-1 Acc (\%)\\
        \midrule
        Large to Base  & 84.2 \\
        Large to Small  & 80.8 \\
        Base to Base & 83.7 \\
        Base  to Small  & 78.6 \\
        \bottomrule
    \end{tabular}
    
    \label{tab:dissize}
    \vskip -0.1in
\end{table}

\textbf{Mask Ratio.} For the mask ratio, we evaluate with different mask proportion. The results are shown in \textbf{Table \ref{tab:disratio} }. According to the results, we can see that as the mask ratio increases, the performance first improves then drops. When mask ratio is low, the learning task is easy and the student does not learn enough information. A higher mask ratio within a reasonable range makes the learning task more challenging and forces the model to extract more information from the data. However, when the mask ratio exceeds some specific threshold, the model losses too much information from the original data and the performance begins to degrade. As we can see, the proposed MOMA adopts much higher mask ratio than the original MAE, enabling faster and more efficient learning.

\begin{table}
    \centering
    \caption{\textbf{Comparison of fine-tuning accuracy on ImageNet-1K for different mask ratio.} All methods use ViT-Base model for both teacher and student, which distill from pre-trained MoCo to pre-trained MAE.}
    \vskip 0.15in
    \begin{tabular}{lr}
        \toprule
        Mask Ratio   & Top-1 Acc (\%)\\
        \midrule
        75\% & 83.4 \\
        80\% & 83.7 \\
        85\%   & 83.8 \\
        90\%   & 84.0 \\
        95\% & 83.6 \\
        \bottomrule
    \end{tabular}
    
    \label{tab:disratio}
    \vskip -0.1in
\end{table}

\textbf{Further Considerations.} By default, we pre-trained our framework for 100 epochs, which is significantly less epochs compared to the existing approaches and computationally efficient. In the experiment, we also tested the extreme case, where we only pre-trained the distillation model for 50 epochs. By halving the training epochs, we saw a performance degradation of around 1.6\%, which is still acceptable. This is critical when computational resource is limited and we need to consider the performance and cost trade-off.

Moreover, we investigated the effect of data augmentation. By default, strong data augmentation operations from MoCo is only applied for the MoCo teacher and MAE student setup. Applying strong data augmentation for MAE teacher is not useful as MAE teacher is never trained on such augmented samples, so it will not yield meaningful representations. We attempted to remove the strong data augmentation for MoCo teacher and observed a performance degradation of 1.3\%. This suggests that effective data augmentation for MoCo Pre-trained teacher is necessary and can boost performance.

Furthermore, we explored the stop gradient operation used in MoCo. By default, we did not apply any stop gradient operation for the patch embedding in both the teacher and student models (unlike MoCo v3, where the patch embedding is frozen) for all our experiments. We attempted to use stop gradient operation for the student in our framework, and this did not lead to a significant change in the performance. The results indicate that stop-gradient / patch frozen is not necessary for our proposed framework.

We also investigated whether the frameworks could be further accelerated by feeding masked inputs into the teacher model. However, masking the teacher branch led to significant performance degradation, compromising the overall performance of the framework. Therefore, it is critical to feed the unmasked image for the teachers to give effective guidance for the student model.

\subsection{Transfer Learning on Downstream Task}

\textbf{Semantic Segmentation.}
We evaluated the proposed framework on downstream segmentation tasks using the ADE20K \cite{zhou2019semantic} dataset with UperNet \cite{xiao2018unified} framework. ADE20K contains 25K images in 150 classes, and we utilized images of size $512 \times 512$. We fine-tuned the pre-trained models (already finetuned on ImageNet-1K) for 160K steps on ADE20K. The pre-trained model is delivered using ViT-Large MAE as the teacher and ViT-Base MoCo as the student. By default, we used a batch size of 16, weight decay of 0.05, the layer decay rate of 0.65, and learning rate of 1e-4. According to the results shown in \textbf{Table \ref{tab:disseg}}, our proposed method outperforms the other methods, presenting great ability in the downstream task.

\begin{table}
    \centering
    \caption{\textbf{Comparison of semantic segmentation performance on ADE20K.} All methods use ViT-Base model. The MoCo v3 entry uses the results in \protect\cite{tao2022siamese}. }
    \vskip 0.15in
    \begin{tabular}{lr}
        \toprule
        Method   & mIoU (\%)\\
        \midrule
        Sup. \cite{liu2021swin} & 46.6 \\
        MAE \cite{he2022masked} & 48.1\\
        MoCo v3 \cite{chen2021empirical}   & 47.3 \\
        MOMA   & 48.6 \\
        \bottomrule
    \end{tabular}
    
    \label{tab:disseg}
    \vskip -0.1in
\end{table}

\textbf{Classification Task.} We explored the transfer learning ability on image classification tasks. We adopted two smaller datasets, CIFAR-10 \cite{krizhevsky2009learning} and CIFAR-100 \cite{krizhevsky2009learning}. Both of them have 60,000 training (50,000 training and 10,000 testing) of size $32 \times 32$. CIFAR-10 has 10 classes and CIFAR-100 has 100 classes. We follow the fine-tuning procedure described in \cite{dosovitskiy2020image} and \cite{chen2021empirical}. The results are shown in \textbf{Table \ref{tab:discls}}. As we can see from the results, the proposed method (i.e., ViT-Large MAE Teacher and ViT-Base MoCo Student) achieves superior or competitive results among all the methods, showing great performance for transfer learning.

\begin{table}
    \centering
    \caption{\textbf{Comparison of transfer learning accuracy (\%) on downstream datasets.} All methods use ViT-Base model. Thd random initialization and supervised learning entries are from \protect\cite{dosovitskiy2020image}}
    \vskip 0.15in
    \begin{tabular}{lrr}
        \toprule
        Method & CIFAR-10   & CIFAR-100\\
        \midrule
        Random Init. &77.8 & 48.5\\
        Sup. & 98.1 & 87.1\\
        MAE  & 99.1 & 91.6\\
        MoCo v3   &98.9 & 90.5\\
        MOMA   & 99.2 & 91.4 \\
        \bottomrule
    \end{tabular}
    
    \label{tab:discls}
    \vskip -0.1in
\end{table}

\section{Conclusion}

Self-supervised learning has achieved exceptional results in various tasks, where contrastive learning and masked image modelling are the two mainstream approaches. However, they usually require high computational resources in addition to their individual limitations. The two methods are complementaty to each other and can be combined effecitvely to leverage their respective strength. To this end, we proposed MOMA, which distills knowledge from self-supervised pre-trained models in a self-supervised manner. MOMA fuses knowldge from contrastive learning and masked image modelling pre-trained models, yielding more powerful, compact and semantically meaningful representations. The proposed method achieves competitive results across different vision datasets and tasks. Additionally, MOMA requires significantly less training epochs compared to the existing self-supervised learning approaches. Furthermore, extremely high mask ratio enables the proposed framework to be fast and efficient, saving computational resource and energy. We hope our work can inspire future studies on how to effectively utilize self-supervised learning in an effective and efficient way.





\bibliography{example_paper}

\begin{thebibliography}{46}
\providecommand{\natexlab}[1]{#1}
\providecommand{\url}[1]{\texttt{#1}}
\expandafter\ifx\csname urlstyle\endcsname\relax
  \providecommand{\doi}[1]{doi: #1}\else
  \providecommand{\doi}{doi: \begingroup \urlstyle{rm}\Url}\fi

\bibitem[Ba et~al.(2016)Ba, Kiros, and Hinton]{ba2016layer}
Ba, J.~L., Kiros, J.~R., and Hinton, G.~E.
\newblock Layer normalization.
\newblock \emph{arXiv preprint arXiv:1607.06450}, 2016.

\bibitem[Bai et~al.(2022)Bai, Wang, Xiao, Wei, Wang, Yuille, Zhou, and
  Xie]{bai2022masked}
Bai, Y., Wang, Z., Xiao, J., Wei, C., Wang, H., Yuille, A., Zhou, Y., and Xie,
  C.
\newblock Masked autoencoders enable efficient knowledge distillers.
\newblock \emph{arXiv preprint arXiv:2208.12256}, 2022.

\bibitem[Bao et~al.(2021)Bao, Dong, and Wei]{bao2021beit}
Bao, H., Dong, L., and Wei, F.
\newblock Beit: Bert pre-training of image transformers.
\newblock \emph{arXiv preprint arXiv:2106.08254}, 2021.

\bibitem[Brown et~al.(2020)Brown, Mann, Ryder, Subbiah, Kaplan, Dhariwal,
  Neelakantan, Shyam, Sastry, Askell, et~al.]{brown2020language}
Brown, T., Mann, B., Ryder, N., Subbiah, M., Kaplan, J.~D., Dhariwal, P.,
  Neelakantan, A., Shyam, P., Sastry, G., Askell, A., et~al.
\newblock Language models are few-shot learners.
\newblock \emph{Advances in neural information processing systems},
  33:\penalty0 1877--1901, 2020.

\bibitem[Caron et~al.(2020)Caron, Misra, Mairal, Goyal, Bojanowski, and
  Joulin]{caron2020unsupervised}
Caron, M., Misra, I., Mairal, J., Goyal, P., Bojanowski, P., and Joulin, A.
\newblock Unsupervised learning of visual features by contrasting cluster
  assignments.
\newblock \emph{Advances in Neural Information Processing Systems},
  33:\penalty0 9912--9924, 2020.

\bibitem[Caron et~al.(2021)Caron, Touvron, Misra, J{\'e}gou, Mairal,
  Bojanowski, and Joulin]{caron2021emerging}
Caron, M., Touvron, H., Misra, I., J{\'e}gou, H., Mairal, J., Bojanowski, P.,
  and Joulin, A.
\newblock Emerging properties in self-supervised vision transformers.
\newblock In \emph{Proceedings of the IEEE/CVF International Conference on
  Computer Vision}, pp.\  9650--9660, 2021.

\bibitem[Chen et~al.(2020{\natexlab{a}})Chen, Radford, Child, Wu, Jun, Luan,
  and Sutskever]{chen2020generative}
Chen, M., Radford, A., Child, R., Wu, J., Jun, H., Luan, D., and Sutskever, I.
\newblock Generative pretraining from pixels.
\newblock In \emph{International conference on machine learning}, pp.\
  1691--1703. PMLR, 2020{\natexlab{a}}.

\bibitem[Chen et~al.(2020{\natexlab{b}})Chen, Kornblith, Norouzi, and
  Hinton]{chen2020simple}
Chen, T., Kornblith, S., Norouzi, M., and Hinton, G.
\newblock A simple framework for contrastive learning of visual
  representations.
\newblock In \emph{International conference on machine learning}, pp.\
  1597--1607. PMLR, 2020{\natexlab{b}}.

\bibitem[Chen et~al.(2020{\natexlab{c}})Chen, Kornblith, Swersky, Norouzi, and
  Hinton]{chen2020big}
Chen, T., Kornblith, S., Swersky, K., Norouzi, M., and Hinton, G.~E.
\newblock Big self-supervised models are strong semi-supervised learners.
\newblock \emph{Advances in neural information processing systems},
  33:\penalty0 22243--22255, 2020{\natexlab{c}}.

\bibitem[Chen \& He(2021)Chen and He]{chen2021exploring}
Chen, X. and He, K.
\newblock Exploring simple siamese representation learning.
\newblock In \emph{Proceedings of the IEEE/CVF Conference on Computer Vision
  and Pattern Recognition}, pp.\  15750--15758, 2021.

\bibitem[Chen et~al.(2020{\natexlab{d}})Chen, Fan, Girshick, and
  He]{chen2020improved}
Chen, X., Fan, H., Girshick, R., and He, K.
\newblock Improved baselines with momentum contrastive learning.
\newblock \emph{arXiv preprint arXiv:2003.04297}, 2020{\natexlab{d}}.

\bibitem[Chen et~al.(2021)Chen, Xie, and He]{chen2021empirical}
Chen, X., Xie, S., and He, K.
\newblock An empirical study of training self-supervised vision transformers.
\newblock In \emph{Proceedings of the IEEE/CVF International Conference on
  Computer Vision}, pp.\  9640--9649, 2021.

\bibitem[Chung et~al.(2021)Chung, Zhang, Han, Chiu, Qin, Pang, and
  Wu]{chung2021w2v}
Chung, Y.-A., Zhang, Y., Han, W., Chiu, C.-C., Qin, J., Pang, R., and Wu, Y.
\newblock W2v-bert: Combining contrastive learning and masked language modeling
  for self-supervised speech pre-training.
\newblock In \emph{2021 IEEE Automatic Speech Recognition and Understanding
  Workshop (ASRU)}, pp.\  244--250. IEEE, 2021.

\bibitem[Cubuk et~al.(2018)Cubuk, Zoph, Mane, Vasudevan, and
  Le]{cubuk2018autoaugment}
Cubuk, E.~D., Zoph, B., Mane, D., Vasudevan, V., and Le, Q.~V.
\newblock Autoaugment: Learning augmentation policies from data.
\newblock \emph{arXiv preprint arXiv:1805.09501}, 2018.

\bibitem[Deng et~al.(2009)Deng, Dong, Socher, Li, Li, and
  Fei-Fei]{deng2009imagenet}
Deng, J., Dong, W., Socher, R., Li, L.-J., Li, K., and Fei-Fei, L.
\newblock Imagenet: A large-scale hierarchical image database.
\newblock In \emph{2009 IEEE conference on computer vision and pattern
  recognition}, pp.\  248--255. Ieee, 2009.

\bibitem[Devlin et~al.(2018)Devlin, Chang, Lee, and Toutanova]{devlin2018bert}
Devlin, J., Chang, M.-W., Lee, K., and Toutanova, K.
\newblock Bert: Pre-training of deep bidirectional transformers for language
  understanding.
\newblock \emph{arXiv preprint arXiv:1810.04805}, 2018.

\bibitem[DeVries \& Taylor(2017)DeVries and Taylor]{devries2017improved}
DeVries, T. and Taylor, G.~W.
\newblock Improved regularization of convolutional neural networks with cutout.
\newblock \emph{arXiv preprint arXiv:1708.04552}, 2017.

\bibitem[Dosovitskiy et~al.(2020)Dosovitskiy, Beyer, Kolesnikov, Weissenborn,
  Zhai, Unterthiner, Dehghani, Minderer, Heigold, Gelly,
  et~al.]{dosovitskiy2020image}
Dosovitskiy, A., Beyer, L., Kolesnikov, A., Weissenborn, D., Zhai, X.,
  Unterthiner, T., Dehghani, M., Minderer, M., Heigold, G., Gelly, S., et~al.
\newblock An image is worth 16x16 words: Transformers for image recognition at
  scale.
\newblock \emph{arXiv preprint arXiv:2010.11929}, 2020.

\bibitem[Grill et~al.(2020)Grill, Strub, Altch{\'e}, Tallec, Richemond,
  Buchatskaya, Doersch, Avila~Pires, Guo, Gheshlaghi~Azar,
  et~al.]{grill2020bootstrap}
Grill, J.-B., Strub, F., Altch{\'e}, F., Tallec, C., Richemond, P.,
  Buchatskaya, E., Doersch, C., Avila~Pires, B., Guo, Z., Gheshlaghi~Azar, M.,
  et~al.
\newblock Bootstrap your own latent-a new approach to self-supervised learning.
\newblock \emph{Advances in neural information processing systems},
  33:\penalty0 21271--21284, 2020.

\bibitem[He et~al.(2020)He, Fan, Wu, Xie, and Girshick]{he2020momentum}
He, K., Fan, H., Wu, Y., Xie, S., and Girshick, R.
\newblock Momentum contrast for unsupervised visual representation learning.
\newblock In \emph{Proceedings of the IEEE/CVF conference on computer vision
  and pattern recognition}, pp.\  9729--9738, 2020.

\bibitem[He et~al.(2022)He, Chen, Xie, Li, Doll{\'a}r, and
  Girshick]{he2022masked}
He, K., Chen, X., Xie, S., Li, Y., Doll{\'a}r, P., and Girshick, R.
\newblock Masked autoencoders are scalable vision learners.
\newblock In \emph{Proceedings of the IEEE/CVF Conference on Computer Vision
  and Pattern Recognition}, pp.\  16000--16009, 2022.

\bibitem[Hinton et~al.(2015)Hinton, Vinyals, Dean,
  et~al.]{hinton2015distilling}
Hinton, G., Vinyals, O., Dean, J., et~al.
\newblock Distilling the knowledge in a neural network.
\newblock \emph{arXiv preprint arXiv:1503.02531}, 2\penalty0 (7), 2015.

\bibitem[Huang et~al.(2022)Huang, Jin, Lu, Hou, Cheng, Fu, Shen, and
  Feng]{huang2022contrastive}
Huang, Z., Jin, X., Lu, C., Hou, Q., Cheng, M.-M., Fu, D., Shen, X., and Feng,
  J.
\newblock Contrastive masked autoencoders are stronger vision learners.
\newblock \emph{arXiv preprint arXiv:2207.13532}, 2022.

\bibitem[Krizhevsky et~al.(2009)Krizhevsky, Hinton,
  et~al.]{krizhevsky2009learning}
Krizhevsky, A., Hinton, G., et~al.
\newblock Learning multiple layers of features from tiny images.
\newblock 2009.

\bibitem[Lin et~al.(2014)Lin, Maire, Belongie, Hays, Perona, Ramanan,
  Doll{\'a}r, and Zitnick]{lin2014microsoft}
Lin, T.-Y., Maire, M., Belongie, S., Hays, J., Perona, P., Ramanan, D.,
  Doll{\'a}r, P., and Zitnick, C.~L.
\newblock Microsoft coco: Common objects in context.
\newblock In \emph{European conference on computer vision}, pp.\  740--755.
  Springer, 2014.

\bibitem[Liu et~al.(2022)Liu, Zhou, Kong, Lin, and Ji]{liu2022exploring}
Liu, X., Zhou, J., Kong, T., Lin, X., and Ji, R.
\newblock Exploring target representations for masked autoencoders.
\newblock \emph{arXiv preprint arXiv:2209.03917}, 2022.

\bibitem[Liu et~al.(2021)Liu, Lin, Cao, Hu, Wei, Zhang, Lin, and
  Guo]{liu2021swin}
Liu, Z., Lin, Y., Cao, Y., Hu, H., Wei, Y., Zhang, Z., Lin, S., and Guo, B.
\newblock Swin transformer: Hierarchical vision transformer using shifted
  windows.
\newblock In \emph{Proceedings of the IEEE/CVF International Conference on
  Computer Vision}, pp.\  10012--10022, 2021.

\bibitem[Loshchilov \& Hutter(2016)Loshchilov and Hutter]{loshchilov2016sgdr}
Loshchilov, I. and Hutter, F.
\newblock Sgdr: Stochastic gradient descent with warm restarts.
\newblock \emph{arXiv preprint arXiv:1608.03983}, 2016.

\bibitem[Loshchilov \& Hutter(2017)Loshchilov and
  Hutter]{loshchilov2017decoupled}
Loshchilov, I. and Hutter, F.
\newblock Decoupled weight decay regularization.
\newblock \emph{arXiv preprint arXiv:1711.05101}, 2017.

\bibitem[Mishra et~al.(2022)Mishra, Robinson, Chang, Jacobs, Sarna, Maschinot,
  and Krishnan]{mishra2022simple}
Mishra, S., Robinson, J., Chang, H., Jacobs, D., Sarna, A., Maschinot, A., and
  Krishnan, D.
\newblock A simple, efficient and scalable contrastive masked autoencoder for
  learning visual representations.
\newblock \emph{arXiv preprint arXiv:2210.16870}, 2022.

\bibitem[Oord et~al.(2018)Oord, Li, and Vinyals]{oord2018representation}
Oord, A. v.~d., Li, Y., and Vinyals, O.
\newblock Representation learning with contrastive predictive coding.
\newblock \emph{arXiv preprint arXiv:1807.03748}, 2018.

\bibitem[Pathak et~al.(2016)Pathak, Krahenbuhl, Donahue, Darrell, and
  Efros]{pathak2016context}
Pathak, D., Krahenbuhl, P., Donahue, J., Darrell, T., and Efros, A.~A.
\newblock Context encoders: Feature learning by inpainting.
\newblock In \emph{Proceedings of the IEEE conference on computer vision and
  pattern recognition}, pp.\  2536--2544, 2016.

\bibitem[Peng et~al.(2022)Peng, Dong, Bao, Ye, and Wei]{peng2022unified}
Peng, Z., Dong, L., Bao, H., Ye, Q., and Wei, F.
\newblock A unified view of masked image modeling.
\newblock \emph{arXiv preprint arXiv:2210.10615}, 2022.

\bibitem[Radford et~al.(2021)Radford, Kim, Hallacy, Ramesh, Goh, Agarwal,
  Sastry, Askell, Mishkin, Clark, et~al.]{radford2021learning}
Radford, A., Kim, J.~W., Hallacy, C., Ramesh, A., Goh, G., Agarwal, S., Sastry,
  G., Askell, A., Mishkin, P., Clark, J., et~al.
\newblock Learning transferable visual models from natural language
  supervision.
\newblock In \emph{International Conference on Machine Learning}, pp.\
  8748--8763. PMLR, 2021.

\bibitem[Robinson et~al.(2020)Robinson, Chuang, Sra, and
  Jegelka]{robinson2020contrastive}
Robinson, J., Chuang, C.-Y., Sra, S., and Jegelka, S.
\newblock Contrastive learning with hard negative samples.
\newblock \emph{arXiv preprint arXiv:2010.04592}, 2020.

\bibitem[Tao et~al.(2022)Tao, Zhu, Huang, Qiao, Wang, and Dai]{tao2022siamese}
Tao, C., Zhu, X., Huang, G., Qiao, Y., Wang, X., and Dai, J.
\newblock Siamese image modeling for self-supervised vision representation
  learning.
\newblock \emph{arXiv preprint arXiv:2206.01204}, 2022.

\bibitem[Touvron et~al.(2021)Touvron, Cord, Douze, Massa, Sablayrolles, and
  J{\'e}gou]{touvron2021training}
Touvron, H., Cord, M., Douze, M., Massa, F., Sablayrolles, A., and J{\'e}gou,
  H.
\newblock Training data-efficient image transformers \& distillation through
  attention.
\newblock In \emph{International Conference on Machine Learning}, pp.\
  10347--10357. PMLR, 2021.

\bibitem[Wei et~al.(2022{\natexlab{a}})Wei, Fan, Xie, Wu, Yuille, and
  Feichtenhofer]{wei2022masked}
Wei, C., Fan, H., Xie, S., Wu, C.-Y., Yuille, A., and Feichtenhofer, C.
\newblock Masked feature prediction for self-supervised visual pre-training.
\newblock In \emph{Proceedings of the IEEE/CVF Conference on Computer Vision
  and Pattern Recognition}, pp.\  14668--14678, 2022{\natexlab{a}}.

\bibitem[Wei et~al.(2022{\natexlab{b}})Wei, Xie, Zhou, Li, and
  Tian]{wei2022mvp}
Wei, L., Xie, L., Zhou, W., Li, H., and Tian, Q.
\newblock Mvp: Multimodality-guided visual pre-training.
\newblock \emph{arXiv preprint arXiv:2203.05175}, 2022{\natexlab{b}}.

\bibitem[Wu et~al.(2018)Wu, Xiong, Yu, and Lin]{wu2018unsupervised}
Wu, Z., Xiong, Y., Yu, S.~X., and Lin, D.
\newblock Unsupervised feature learning via non-parametric instance
  discrimination.
\newblock In \emph{Proceedings of the IEEE conference on computer vision and
  pattern recognition}, pp.\  3733--3742, 2018.

\bibitem[Xiao et~al.(2018)Xiao, Liu, Zhou, Jiang, and Sun]{xiao2018unified}
Xiao, T., Liu, Y., Zhou, B., Jiang, Y., and Sun, J.
\newblock Unified perceptual parsing for scene understanding.
\newblock In \emph{Proceedings of the European conference on computer vision
  (ECCV)}, pp.\  418--434, 2018.

\bibitem[Xie et~al.(2022)Xie, Zhang, Cao, Lin, Bao, Yao, Dai, and
  Hu]{xie2022simmim}
Xie, Z., Zhang, Z., Cao, Y., Lin, Y., Bao, J., Yao, Z., Dai, Q., and Hu, H.
\newblock Simmim: A simple framework for masked image modeling.
\newblock In \emph{Proceedings of the IEEE/CVF Conference on Computer Vision
  and Pattern Recognition}, pp.\  9653--9663, 2022.

\bibitem[Yao et~al.(2022)Yao, Desai, and Palaniswami]{yao2022masked}
Yao, Y., Desai, N., and Palaniswami, M.
\newblock Masked contrastive representation learning.
\newblock \emph{arXiv preprint arXiv:2211.06012}, 2022.

\bibitem[Zhou et~al.(2019)Zhou, Zhao, Puig, Xiao, Fidler, Barriuso, and
  Torralba]{zhou2019semantic}
Zhou, B., Zhao, H., Puig, X., Xiao, T., Fidler, S., Barriuso, A., and Torralba,
  A.
\newblock Semantic understanding of scenes through the ade20k dataset.
\newblock \emph{International Journal of Computer Vision}, 127\penalty0
  (3):\penalty0 302--321, 2019.

\bibitem[Zhou et~al.(2021)Zhou, Wei, Wang, Shen, Xie, Yuille, and
  Kong]{zhou2021ibot}
Zhou, J., Wei, C., Wang, H., Shen, W., Xie, C., Yuille, A., and Kong, T.
\newblock ibot: Image bert pre-training with online tokenizer.
\newblock \emph{arXiv preprint arXiv:2111.07832}, 2021.

\bibitem[Zhou et~al.(2022)Zhou, Yu, Luo, Wang, and Li]{zhou2022mimco}
Zhou, Q., Yu, C., Luo, H., Wang, Z., and Li, H.
\newblock Mimco: Masked image modeling pre-training with contrastive teacher.
\newblock In \emph{Proceedings of the 30th ACM International Conference on
  Multimedia}, pp.\  4487--4495, 2022.

\end{thebibliography}
\bibliographystyle{icml2023}



\end{document}